\documentclass[conference]{IEEEtran}
\IEEEoverridecommandlockouts

\usepackage{cite}
\usepackage{amsmath,amssymb,amsfonts}
\usepackage{algorithmic}
\usepackage{graphicx}
\usepackage{textcomp}
\usepackage{xcolor}
\usepackage{subcaption}                  
\usepackage{caption}                     
\usepackage{hyperref}  
\usepackage{array} 
\usepackage{titlesec}
\usepackage{afterpage}
\usepackage{lineno}
\usepackage{array} 
\usepackage{adjustbox}
\usepackage{multirow} 
\usepackage{bigstrut} 

\newcolumntype{P}[1]{>{\centering\arraybackslash}p{#1}}

\def\BibTeX{{\rm B\kern-.05em{\sc i\kern-.025em b}\kern-.08em
    T\kern-.1667em\lower.7ex\hbox{E}\kern-.125emX}}
\begin{document}

\title{Shape Proportions and Sphericity in $n$ Dimensions}

\author{\IEEEauthorblockN{ William Franz Lamberti}
\IEEEauthorblockA{\textit{Center for Public Health Genomics} \\
\textit{Department of Biomedical Engineering}\\
\textit{School of Data Science}\\
\textit{University of Virginia}\\
Charlottesville, VA, United States \\
william.f.lamberti@virginia.edu\\
williamfranzlamberti@gmail.com
}
}

\maketitle
\IEEEpubidadjcol

\begin{abstract}
Shape metrics for objects in high dimensions remain sparse.  Those that do exist, such as hyper-volume, remain limited to objects that are better understood such as Platonic solids and $n$-Cubes.  Further, understanding objects of ill-defined shapes in higher dimensions is ambiguous at best.  Past work does not provide a single number to give a qualitative understanding of an object.  For example, the eigenvalues from principal component analysis results in $n$ metrics to describe the shape of an object.  Therefore, we need a single number which can discriminate objects with different shape from one another.   Previous work has developed shape metrics for specific dimensions such as two or three dimensions.  However, there is an opportunity to develop metrics for any desired dimension.  To that end, we present two new shape metrics for objects in a given number of dimensions: hyper-Sphericity and hyper-Shape Proportion (SP).  We explore the proprieties of these metrics on a number of different shapes including $n$-balls.  We then connect these metrics to applications of analyzing the shape of multidimensional data such as the popular Iris dataset.  
\end{abstract}


\section{Introduction}

Objects in hyperdimensional spaces are an active area of research.  Some aspects of these types of objects are well understood such as the volume of $n-$balls (hyperspheres) or $n-$cubes.  However, our understanding of these objects becomes less certain when the type of object in question deviates from the well-studied types of objects.  Further, formal analysis of data as hyper-dimensional images and characterizing its shape is underdeveloped.   

One way to understand an object is by using shape metrics to describe the object in question.  There are numerous different kinds of shape metrics to describe objects in spaces where $n=2$\cite{lamberti_algorithms_2020, euclid_euclids_1728, klinkenberg_review_1994, lopes_fractal_2009, morency_fractal_2003, plotze_leaf_2005, costa_shape_2018,  rosenfeld_compact_1974, kinser_image_2018, harris_combined_1988, hu_visual_1962, flusser_affine_1994, gonzalez_digital_2009}.  However, the number of metrics that exist for large $n$ are limited.  Further, translating how to compute these metrics for any type of shape in image data is nontrivial.  

Eigenvalues can be used as a set of shape metrics to describe the shape of data\cite{kinser_image_2018}.  Eigenvalues correspond to the relative length of each axis of the data\cite{lamberti_william_franz_overview_2022}.  However, this corresponds to $n$ values to describe data in $n$ dimensions.  Thus, there is no singular value that can describe the shape of data in large multidimensional spaces.  This large number of shape metrics becomes unwieldy and cumbersome when $n$ is large.  Thus, a single number to describe any $n$D shape or data would provide valuable insight to the nature of the object or data. 

To help alleviate these issues, we provide two shape metrics to describe any given object in $n-$ dimensional ($n$D) space: hyper-shape proportions and hyper-sphericity.  While both of these shape metrics have already been described for their 2D and 3D counterparts \cite{lamberti_extracting_2022, wadell_volume_1935}, we extend their applications to any given $n$ such that $n\in \{2,3,4...\} $.  We prove that hyper-SP has unique values for platonic solids, $n$-simplexes, $n$-cubes, $n-$orthoplexes, and $n$-balls.  We also prove that hyper-sphericity has a unique value for $n$-balls. 

We present an approach for computing these metrics for objects with ill-defined shapes by converting the objects into hyperdimensional images.  While similar approaches have been done to describe 2D shapes\cite{lamberti_using_2022}, this is one of the first methods to describe large $n$ dimensional objects.  For example, describing data using these shape metrics can provide insights to how objects exist in these spaces.  By converting sample data from a  multidimensional Normal to a 3D image, we provide an example of our computational approach when $n=3$.  This was explored in depth for $n$-balls across several dimensions.  This approach has wide applicability in data rich fields to describe multidimensional data.  Our code is provided on our GitHub: \url{https://github.com/billyl320/ndim_shape_metrics}.


\section{Shape Metrics}
Both shape metrics provide insights to how objects exist relative to an $n-$ball.  SP is primarily concerned with the volumes of a given object, while sphericity is concerned with the volume and surface area (which is sometimes referred to as hyper-surface area or hyper-perimeter).  For notation purposes, let $V$ be the volume of a given object, $V_s$ be the volume of a sphere, and $n$ be the number of dimensions an objects resides within.  

\subsection{SP}
\begin{figure}[h]
    \centering
    \begin{subfigure}[t]{.25\textwidth}
    \centering
    \includegraphics[width=0.70\linewidth]{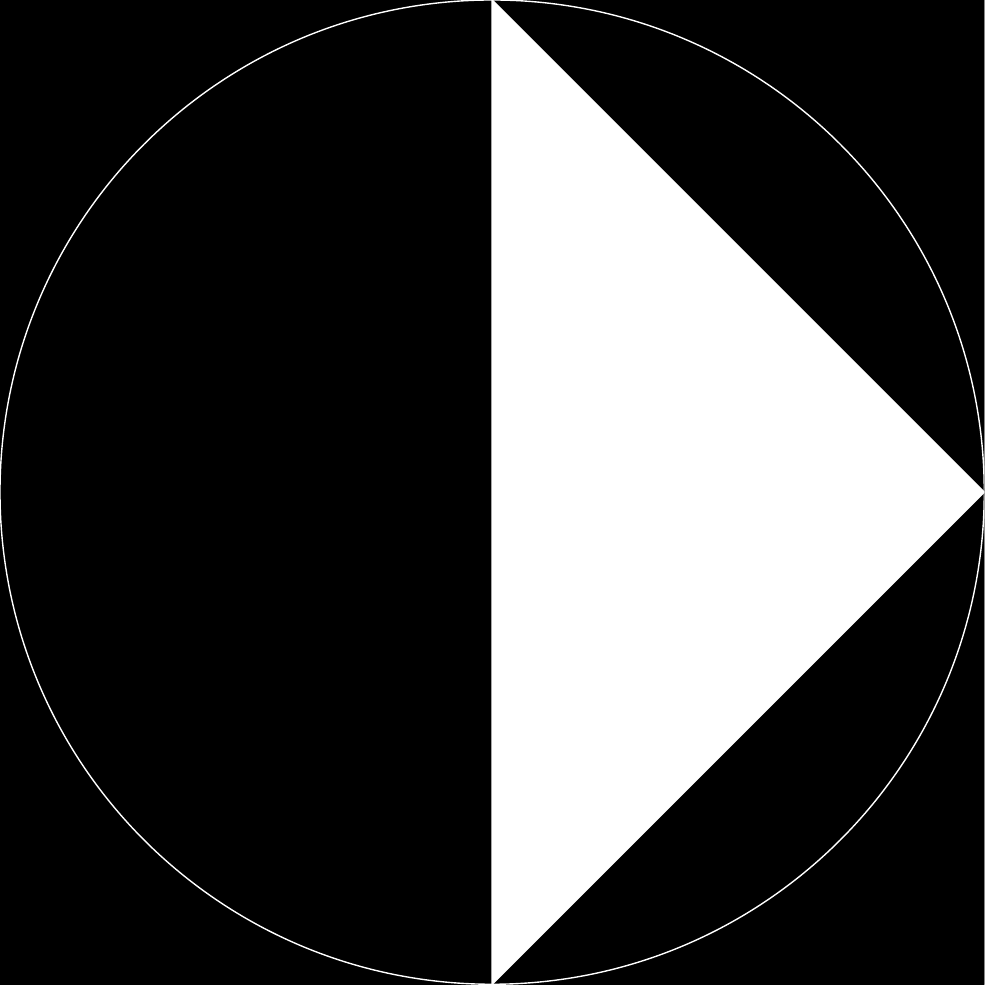}
    \caption{}
    \label{fig: sp_1}
    \end{subfigure}
    \begin{subfigure}[t]{.25\textwidth}
    \centering
    \includegraphics[width=0.70\linewidth]{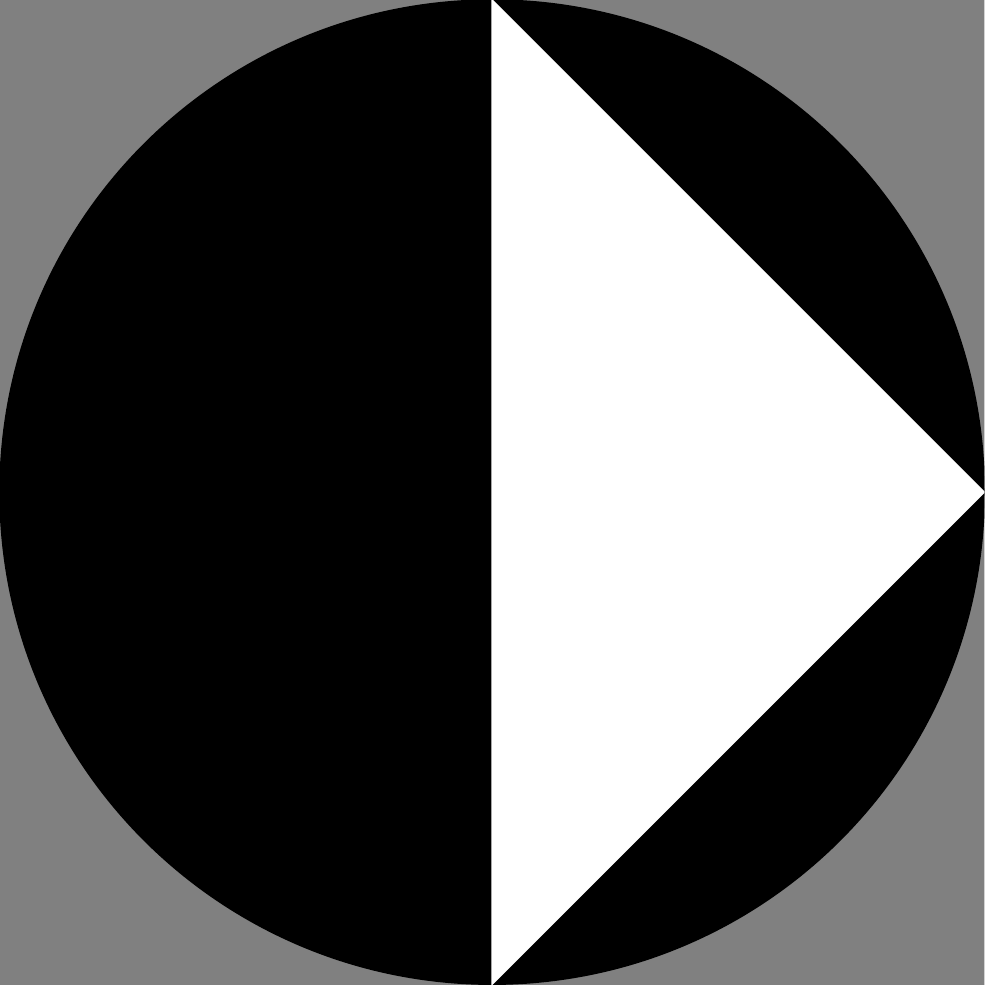}
    \caption{}
    \label{fig: sp_2}
    \end{subfigure}
    \caption{(a) Example with triangle in the minimum encompassing square which encompasses the minimum encompassing circle. (b) Our proposed formulation only considers the circle and triangle.  The grayed section is not included in this different formulation.}
    \label{fig:sp_square}
\end{figure}

SP has already be investigated and proposed for 2D objects\cite{lamberti_algorithms_2020} and has been extended to 3D objects \cite{lamberti_extracting_2022}.  In summary, the original formulation of the SP value for a given object represented the proportion of area that the object occupied relative to the area of the minimum encompassing square which encompassed the minimum encompassing circle.  Figure \ref{fig: sp_1} provides a representation of this formulation.  However, it was noted that SP could be defined by any desired encompassing object.  

We define SP, $p$,  to be
\begin{equation}\label{eq: sp}
    p = \frac{V}{V_s}, \forall n \in \{2,3,4,...\}.
\end{equation}

Thus, $p$ represents the proportion of the volume of an object occupies relative to the object's minimum encompassing $n$-ball's volume.  This formulation is exemplified in Figure \ref{fig: sp_2} Thus, $p\in (0, 1]$.  A value of 1 indicates that the object is a perfect $n$-ball, while a value of 0 indicates that the object has no volume.  Specific kinds of objects will be discussed and proven to have unique values for any given $n$ as follows.  

\subsubsection{Platonic Solids}

For platonic solids, let $V_p$ be the volume of a given platonic solid for $n=3$ and $r$ be the radius of the minimum encompassing sphere.  Thus, SP for a platonic solid is

\begin{equation}
    p = \frac{V_p}{V_s} = \frac{fs \sin{\frac{360^\circ}{s}} \Gamma(\frac{5}{2}) }{6\pi^{\frac{3}{2}}}
\end{equation}

\noindent where $s$ is the number of sides (or edges) and $f$ is the number of faces.  

\textbf{Proof:} 

\begin{equation}
    p = \frac{V}{V_s}
\end{equation}

\begin{equation}
    = \frac{\frac{1}{6}fsr^3\sin{\frac{360^\circ}{s}} }{\frac{r^3\pi^{\frac{3}{2}} } {\Gamma(\frac{5}{2})}}
\end{equation}

\begin{equation}
    = \frac{fs \sin{\frac{360^\circ}{s}} \Gamma(\frac{5}{2}) }{6\pi^{\frac{3}{2}}}\ \ \ \blacksquare
\end{equation}

\subsubsection{$n-$Simplex}

For an $n-$simplex, let $V_s$ be the volume of a given $n-$simplex for a given $n$.  Thus, SP for a $n-$simplex is

\begin{equation}
    p = \frac{\sqrt{n+1} \Gamma(\frac{n}{2} +1 ) 4^{n} }{ n!(12\pi)^{\frac{n}{2}}  }.
\end{equation}

\textbf{Proof:}

\begin{equation}
    p = \frac{V_t}{V_s} 
\end{equation}

\begin{equation}
    = \frac{ \frac{\sqrt{n+1}}{n!\sqrt{2^n}}l^n }{ \frac{r^n\pi^{\frac{n}{2}}}{\Gamma(\frac{n}{2} +1) } }
\end{equation}

\begin{equation}
    = \frac{\sqrt{n+1} \Gamma(\frac{n}{2} +1) }{ n!(2\pi)^{\frac{n}{2}} }\times \frac{ \frac{4^n}{6^{\frac{n}{2}}} r^n}{r^n}
\end{equation}

\begin{equation}
    = \frac{\sqrt{n+1} \Gamma(\frac{n}{2} +1 ) 4^{n} }{ n!(12\pi)^{\frac{n}{2}}  } \ \ \ \blacksquare
\end{equation}

\subsubsection{$n-$Cube}

For an $n-$cube, let $V_c$ be the volume of a given $n-$cube for a given $n$.  Thus, SP for a $n-$cube is

\begin{equation}
    p = \frac{ \Gamma(\frac{n}{2} +1 ) }{ (2\pi)^{\frac{n}{2}}  }
\end{equation}

\textbf{Proof:}

\begin{equation}
    p = \frac{V_c}{V_s} 
\end{equation}

\begin{equation}
    = \frac{ l^n }{ \frac{r^n\pi^{\frac{n}{2}}}{\Gamma(\frac{n}{2} +1) } }
\end{equation}

\begin{equation}
    = \frac{ \Gamma(\frac{n}{2} +1) (\frac{r}{\sqrt{2}})^n }{ r^n\pi^{\frac{n}{2}} }
\end{equation}

\begin{equation}
    = \frac{ \Gamma(\frac{n}{2} +1 ) }{ (2\pi)^{\frac{n}{2}}  } \ \ \ \blacksquare
\end{equation}

\subsubsection{$n-$Orthoplex}

For an $n-$orthoplex, let $V_o$ be the volume of a given $n-$orthoplex for a given $n$.  Thus, SP for a $n-$orthoplex is

\begin{equation}
    p = \frac{ 2^n\Gamma(\frac{n}{2} +1 ) }{ n!\pi^{\frac{n}{2}}  }
\end{equation}

\textbf{Proof:}

\begin{equation}
    p = \frac{V_o}{V_s} 
\end{equation}

\begin{equation}
    = \frac{ \frac{\sqrt{2^n}}{n!}l^n }{ \frac{r^n\pi^{\frac{n}{2}}}{\Gamma(\frac{n}{2} +1) } }
\end{equation}

\begin{equation}
    = \frac{ \sqrt{2^n}\Gamma(\frac{n}{2} +1) (\frac{r}{\sqrt{2}})^n }{ n!\pi^{\frac{n}{2}}r^n }\times \frac{(2r)^n}{2^{\frac{n}{2}}}
\end{equation}

\begin{equation}
    = \frac{ \Gamma(\frac{n}{2} +1 ) }{ (2\pi)^{\frac{n}{2}}  }
\end{equation}

\begin{equation}
    = \frac{ 2^n\Gamma(\frac{n}{2} +1 ) }{ n!\pi^{\frac{n}{2}}  } \ \ \ \blacksquare
\end{equation}

\subsection{Sphericity}
Sphericity is based on the measures of 2D objects called circularity \cite{rosenfeld_compact_1974, kinser_image_2018}.  It has also be extended to 3D objects for a metric with the same name (sphericity) \cite{wadell_volume_1935}.  These two metrics measure how circular or spherical a 2D or 3D object is, respectively.  However, both of these metrics are limited to their respective dimensions.  Thus, our generalized sphericity metric provides a measure $\forall n \in \{2,3,4,...\}$.  

We define sphericity, $\gamma$, to be

\begin{equation}\label{eq: spher}
    \gamma = \frac{nV}{rS}, \forall n \in \{2,3,4,...\}
\end{equation}

\noindent where $S$ is the hyperdimensional surface area of the object.  This value turns out to be 1 for $n-$balls (as we will show).  Thus, values close to 1 indicate objects that are spherical in shape.  Values far away from 1 indicate less spherical objects.

We will now prove that $n-$balls have a sphericity value of 1.  

\textbf{Proof:}

\begin{equation}
    \gamma = \frac{nV}{rS}
\end{equation}

\begin{equation}
    = \frac{n(\pi^{\frac{n}{2}}\times r^2) } {\Gamma(\frac{n}{2}+1)} \div \frac{2\pi^{\frac{n}{2}}\times r^{n-1}\times r }{\Gamma(\frac{n}{2})}
\end{equation}

\begin{equation}
    = \frac{n}{2}\times \frac{\Gamma(\frac{n}{2})}{\Gamma(\frac{n}{2}+1)}
\end{equation}

\begin{equation}
    = \frac{n}{2}\times \frac{\Gamma(\frac{n}{2}+1)\div \frac{n}{2}}{\Gamma(\frac{n}{2}+1)} = 1 \ \ \ \blacksquare
\end{equation}

\section{Applications to Data as Images}\label{sec:res}

\begin{figure}[h]
    \centering
    \begin{subfigure}[t]{.50\textwidth}
    \centering
    \includegraphics[width=0.90\linewidth]{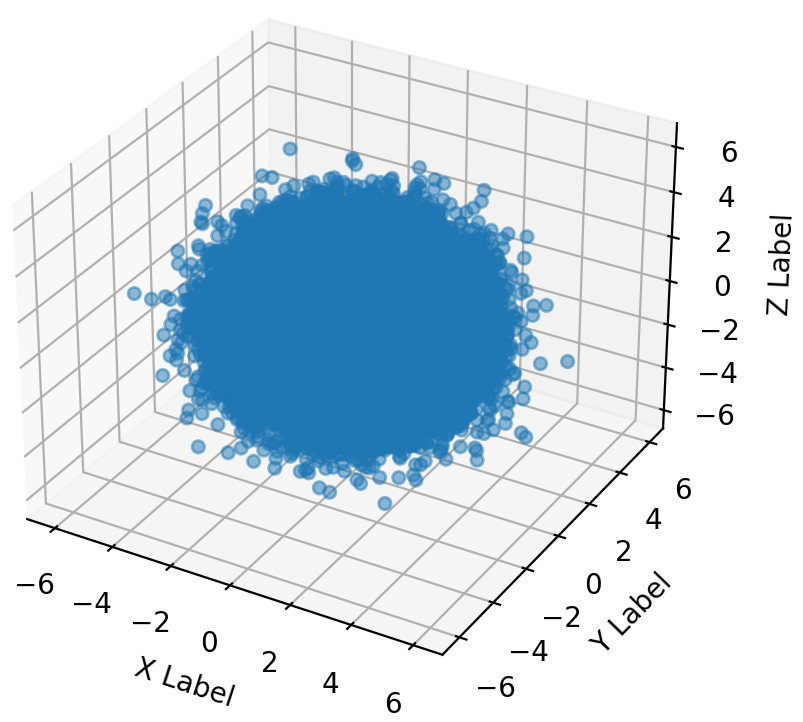}
    \caption{}
    \label{fig: 3d_norm_raw}
    \end{subfigure}\\
    \begin{subfigure}[t]{.50\textwidth}
    \centering
    \includegraphics[width=0.90\linewidth]{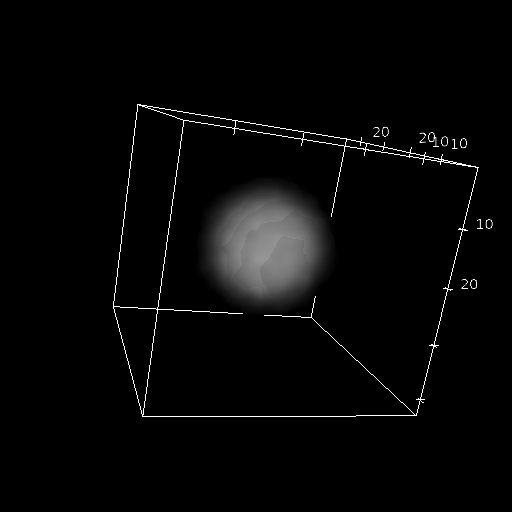}
    \caption{}
    \label{fig: 3d_norm_image}
    \end{subfigure}
    \caption{(a) Example random Normal raw data in 3D space. (b) Resulting random Normal image data in 3D space visualized using ImageJ\cite{tiago_imagej_2012}.}
    \label{fig: image_example}
\end{figure}

As stated previously, the application of this work extends the previous work by converting 2D data \cite{lamberti_using_2022} and 3D data \cite{lamberti_extracting_2022} into $n$D images.  The approach entails converting the feature space into a discretized version using $n$D histograms. Describing this using image operator notation\cite{kinser_image_2018}, $n$D raw data are converted into $n$D images using  \begin{equation}\label{eq: dat2img}
    \textbf{b}[\vec{x}] = \Gamma_{>0}\mathbb{H}_{n} X,  
\end{equation}
\noindent  where $X$ is the input data, $\mathbb{H}_{n}$ converts the data into a $n$D histogram \cite{noauthor_numpyhistogram2d_nodate}, $\Gamma_{>0}$ is the threshold image operator, and $\textbf{b}[\vec{x}]$ is the resulting $n$D image.  Figure \ref{fig: image_example} provides an example of the raw 3D Normal input  data and the resulting image after $\mathbb{H}_{n}$ is applied.

After applying Equation \ref{eq: dat2img}, we would collect various shape metrics of interest.  Calculating the numerator of SP is straightforward.  It is merely\begin{equation}\label{eq: num}
    V = \sum \textbf{b}[\vec{x}]
\end{equation}
where $\sum$ is the summation operator.  This provides the volume of our object.  Calculating the denominator of SP requires multiple steps with the eventual goal of extracting $r$, the minimum encompassing radius of our object.  

\begin{equation}
\label{eq: center}
\vec{c} = \boxtimes \textbf{b}[\vec{x}],
\end{equation}

\begin{equation}
\label{eq: radius}
r = \uparrow \bigvee ((\mathcal{D}_E \textbf{o}[\vec{p}]) \times \textbf{b}[\vec{p}]), 
\end{equation}

\noindent where $\boxtimes$ is the center of mass operator, $\uparrow$ is the ceiling or rounding up math operator,$\mathcal{D}_E$ returns the distance transformation image using Euclidean distance, and $\textbf{o}[\vec{p}]$ is a matrix of 1s except at $\vec{c}$ where the value is 0 and the shape is the same as $\textbf{b}[\vec{p}]$.  Finally, we perform\begin{equation}\label{eq: denom}
    V_s = \frac{\pi^{n/2} r^n}{\Gamma(\frac{n}{2}+1)}
\end{equation}

\noindent where $\Gamma$ is the mathematical gamma function.  Using equations \ref{eq: num} and \ref{eq: denom} allows us to compute Equation \ref{eq: sp}.  

Calculating hyper-sphericity is straightforward.  The numerator of Equation \ref{eq: spher} is merely\begin{equation}\label{eq: num_spher}
    nV = n\times  \sum \textbf{b}[\vec{x}].
\end{equation}

\noindent The denominator is then\begin{equation}\label{eq: denom_spher}
    rS = r \times\sum {\LARGE{(}}\sum \textbf{b}[\vec{x}] - \triangleright_1 \sum \textbf{b}[\vec{x}]{\LARGE{)}}
\end{equation}

\noindent where $\triangleright_1$ is the erosion operator with 1 iteration.  Thus, hyper-sphericity is merely Equation \ref{eq: num_spher} divided by \ref{eq: denom_spher}.  From here, we will present experiments which capture shape metrics on simulated $n$-balls and the Iris dataset\cite{edgar_irises_1935}.  

\subsection{Experimental Setup}

We first simulated $n$D balls \cite{voelker_eciently_2017, harman_decompositional_2010} to evaluate our implementations of hyper-sphericity and hyper-SP across different dimensions and number of bins.  We then evaluated our approach on different bootstrapped subsets of the public Iris dataset\cite{edgar_irises_1935} to evaluate if the shape changes in substantial ways using different bin sizes.  

\subsection{$n$D-Balls}

\begin{figure}[h]
    \centering
    \begin{subfigure}[t]{.25\textwidth}
    \centering
    \includegraphics[width=0.90\linewidth]{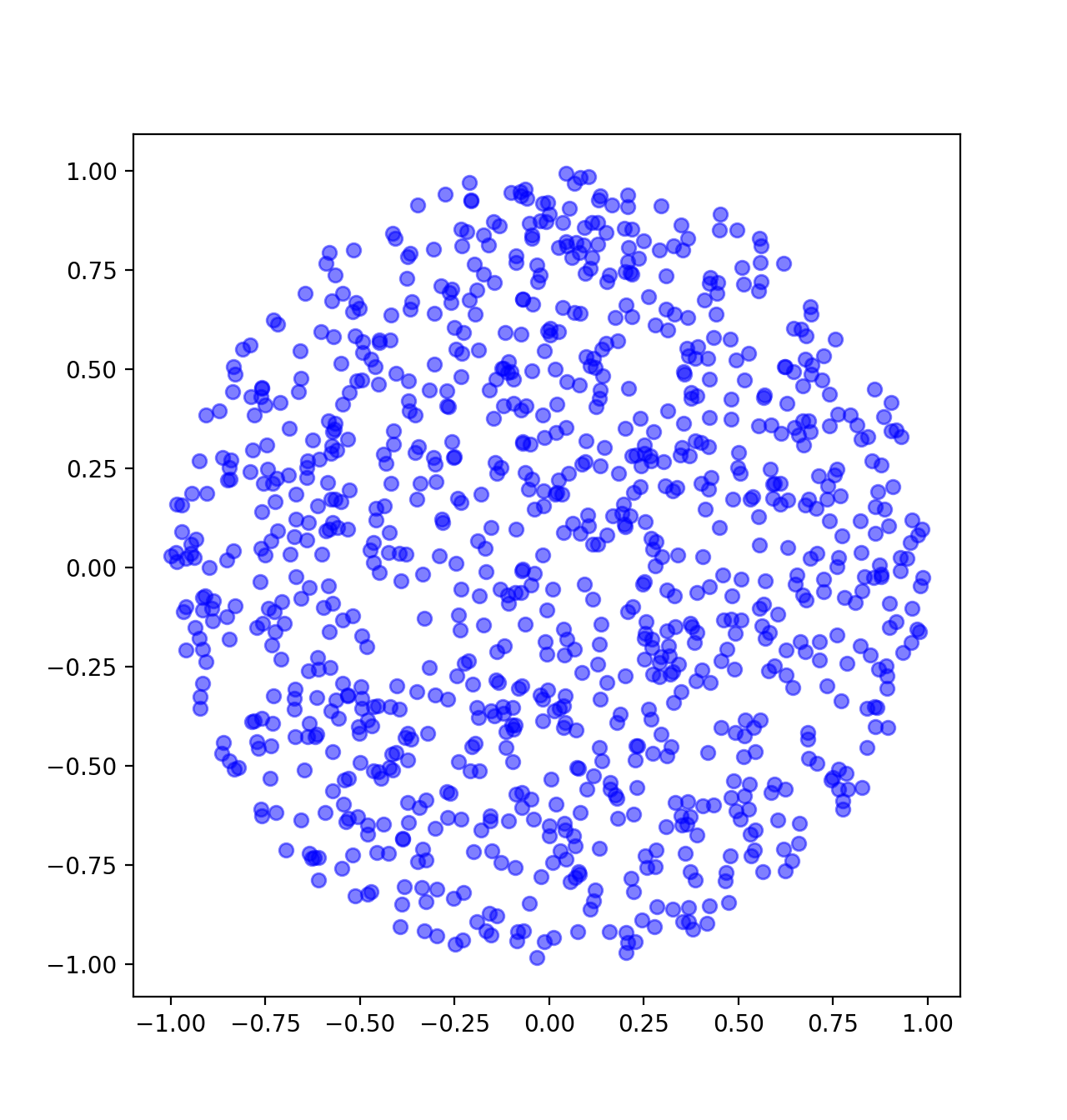}
    \caption{}
    \label{fig: uni_1}
    \end{subfigure}
    \begin{subfigure}[t]{.25\textwidth}
    \centering
    \includegraphics[width=0.90\linewidth]{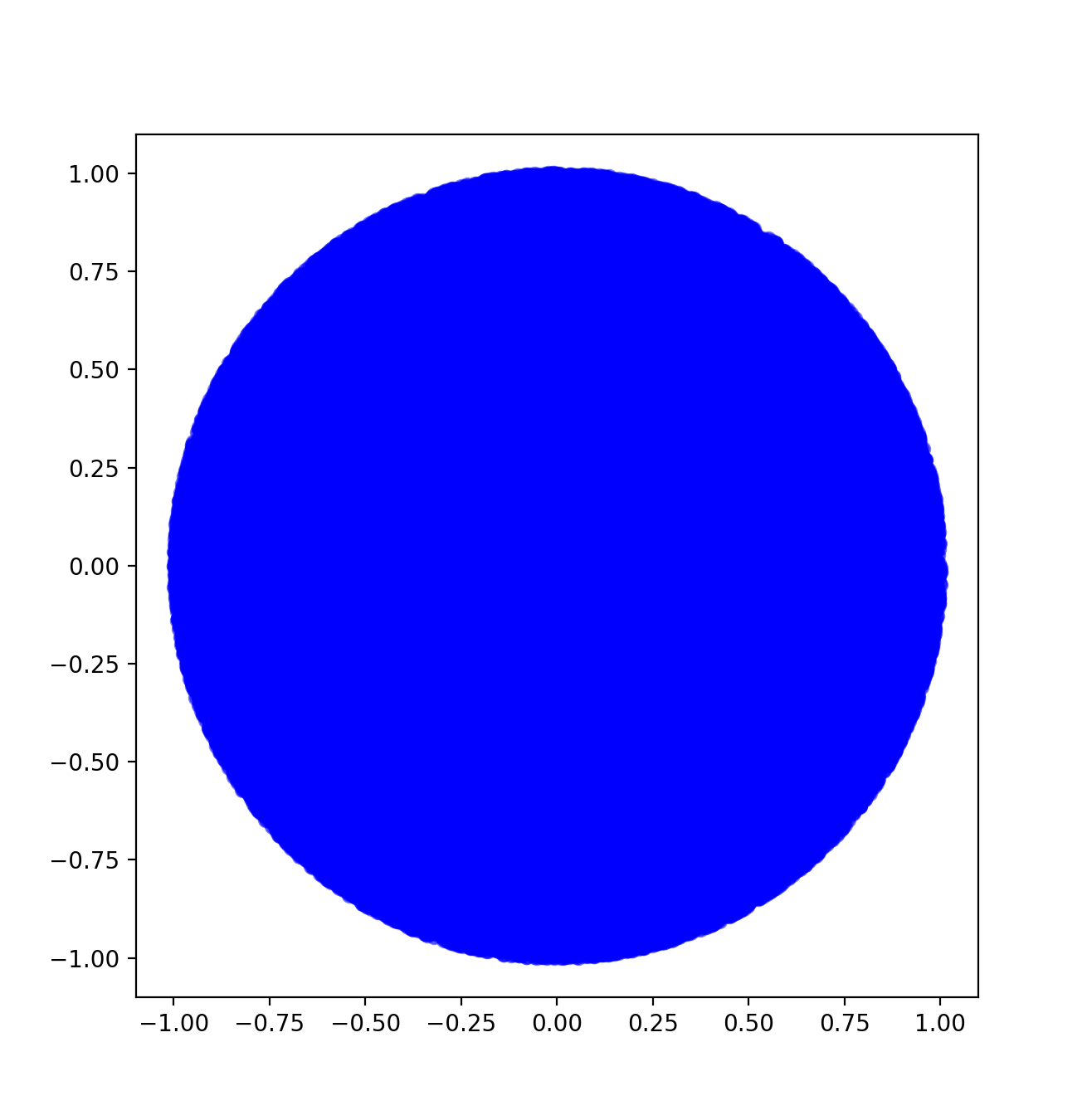}
    \caption{}
    \label{fig: uni_2}
    \end{subfigure}
    \caption{Examples of $nD$ ball dropping algorithm where $n=2$. The number of points for each image is (a) 1,000 and (b) 100,000.}
    \label{fig: uni}
\end{figure}

We first implemented an $n$D ball dropping algorithm \cite{voelker_eciently_2017, harman_decompositional_2010} in Python 3.6\cite{noauthor_python_nodate} using the numpy, random, imageio, scipy.special and scipy.ndimage modules.  By creating sample 2D images, we determined that an appropriate number of points to consider was 100,000 for our purposes(Fig. \ref{fig: uni}).  Next, we tested the number of bins from a range of 4 to 14, using 100 simulated balls per each bin size.  We extracted hyper-SP and hyper-sphericity metrics for each of the 100 samples.  

\begin{figure*}[h]
    \centering
    \begin{subfigure}[t]{.60\textwidth}
    \centering
    \includegraphics[width=0.90\linewidth]{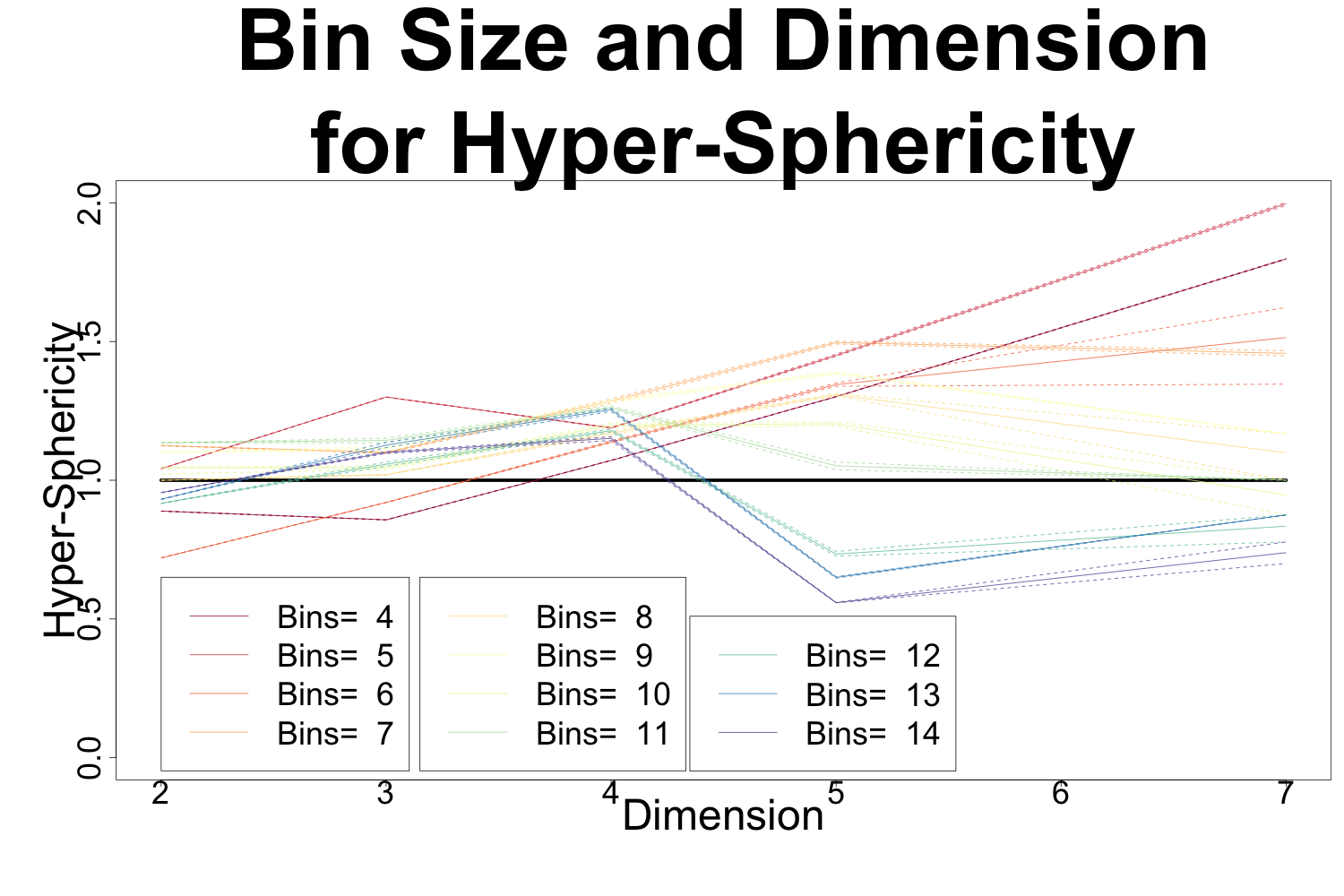}
    \caption{}
    \label{fig: hyper_spher}
    \end{subfigure}\\
    \begin{subfigure}[t]{.60\textwidth}
    \centering
    \includegraphics[width=0.90\linewidth]{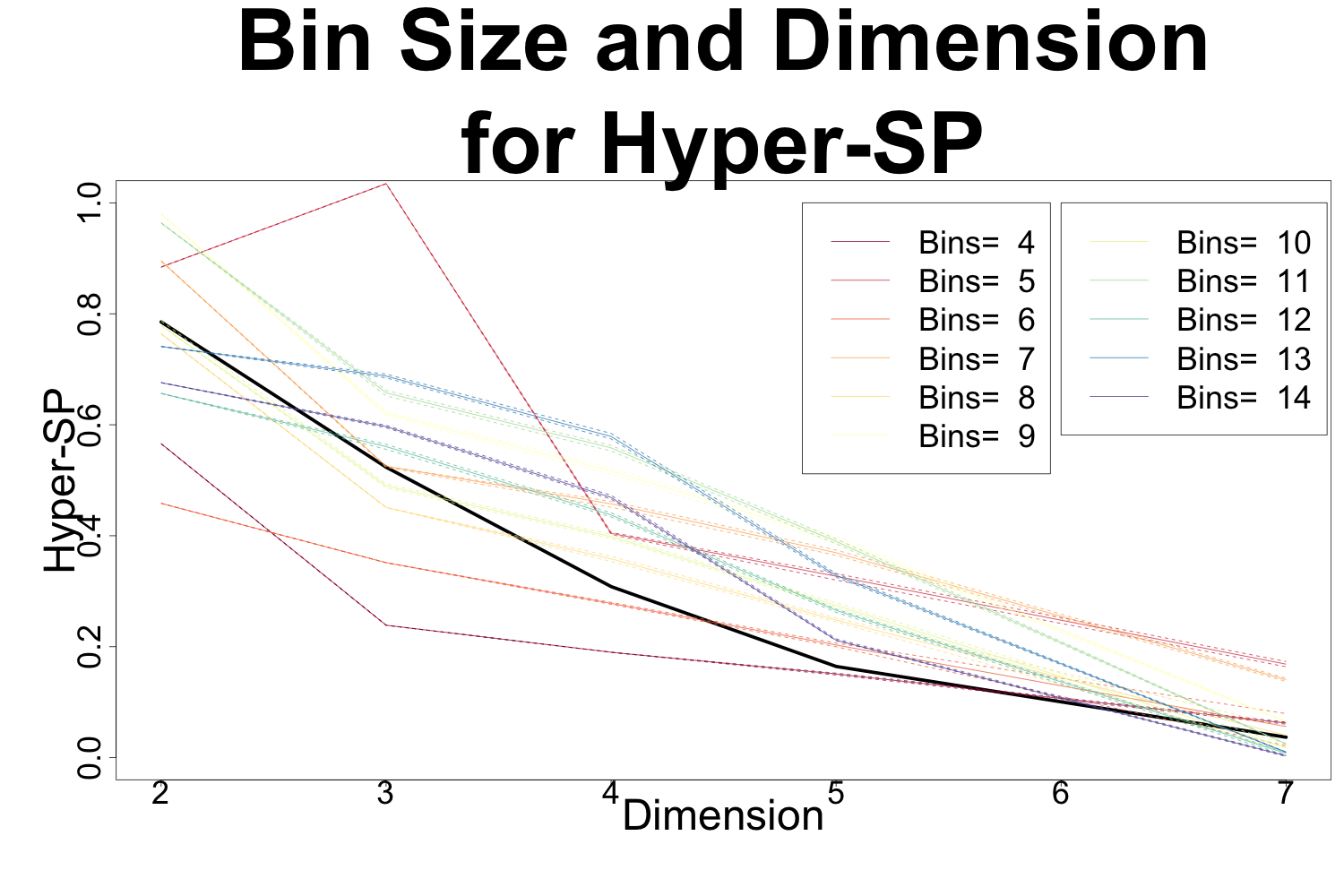}
    \caption{}
    \label{fig: hyper_sp}
    \end{subfigure}\\
    \begin{subfigure}[t]{.60\textwidth}
    \centering
    \includegraphics[width=0.90\linewidth]{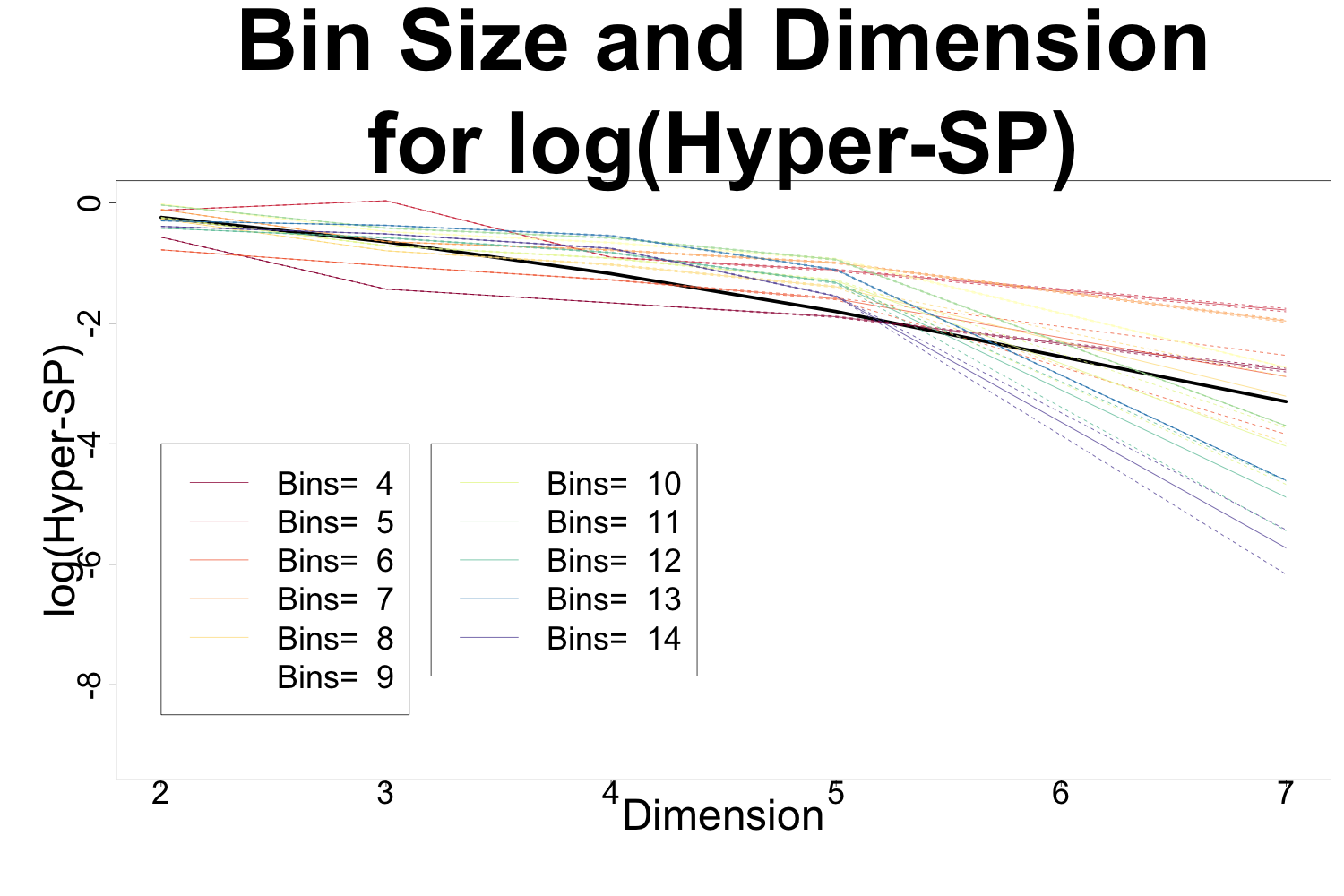}
    \caption{}
    \label{fig: log_hyper_sp}
    \end{subfigure}
    \caption{Estimated values when using different bin sizes across different dimensions for \textbf{(a)} Hyper-Sphericity, \textbf{(b)} Hyper-SP, and the \textbf{(c)} Natural log of Hyper-SP.  The true theoretical value is presented as a black solid line.  The mean estaimted value is presented as a colorized solid line, which is surrounded by the 95\% quantile interval as dotted colorized lines.}
    \label{fig: uniform}
\end{figure*}

Then using R, we calculated the mean and 2.5\% and 97.5\% quantiles of the shape metrics across the different bins.  We then compared these ranges against the theoretical true value across different dimensions (Fig. \ref{fig: uniform}).  

\subsection{Iris Data}

\begin{figure*}[h]
    \centering
    \begin{subfigure}[t]{.90\textwidth}
    \centering
    \includegraphics[width=0.90\linewidth]{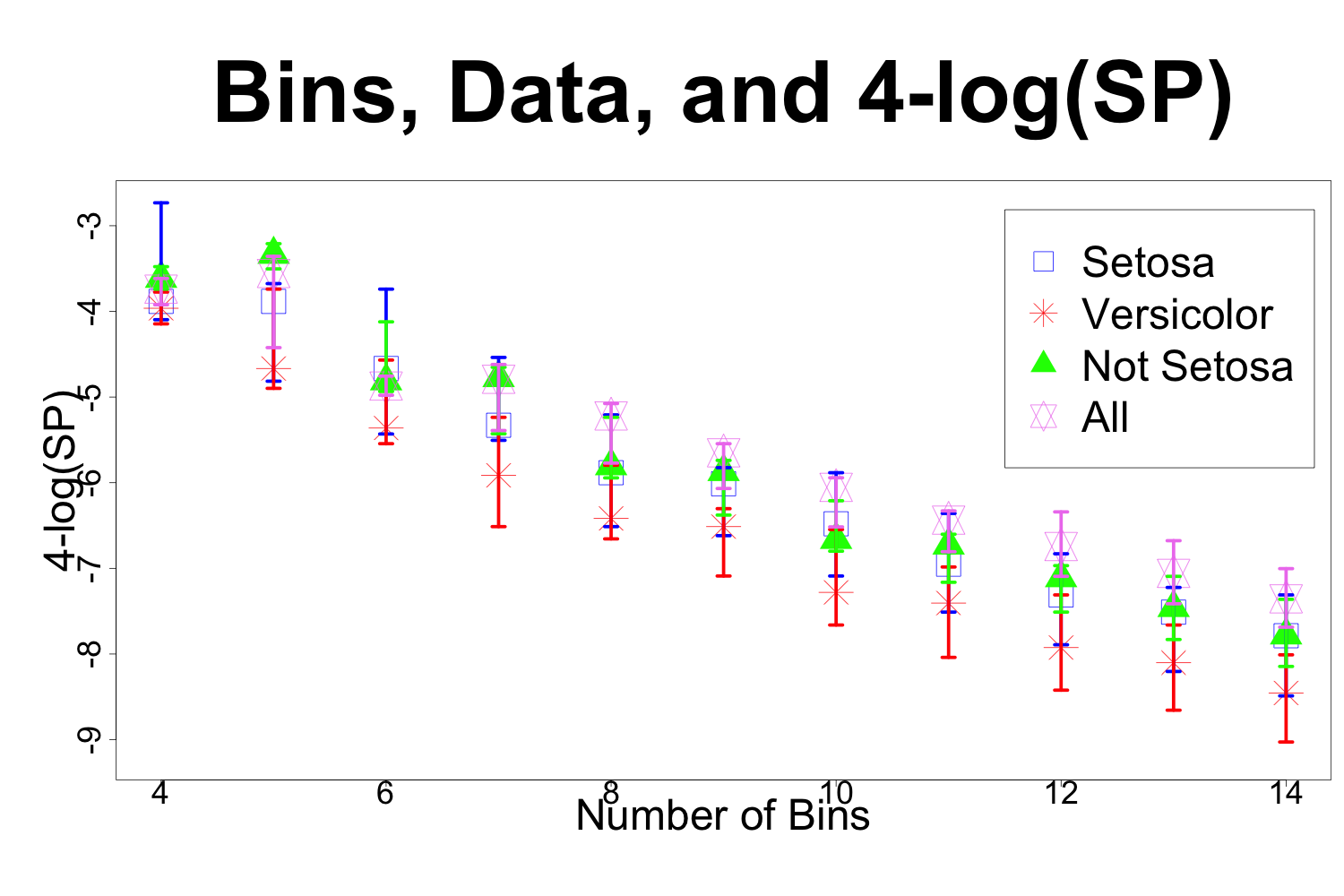}
    \caption{}
    \label{fig: iris_log_sp}
    \end{subfigure}\\
    \begin{subfigure}[t]{.90\textwidth}
    \centering
    \includegraphics[width=0.90\linewidth]{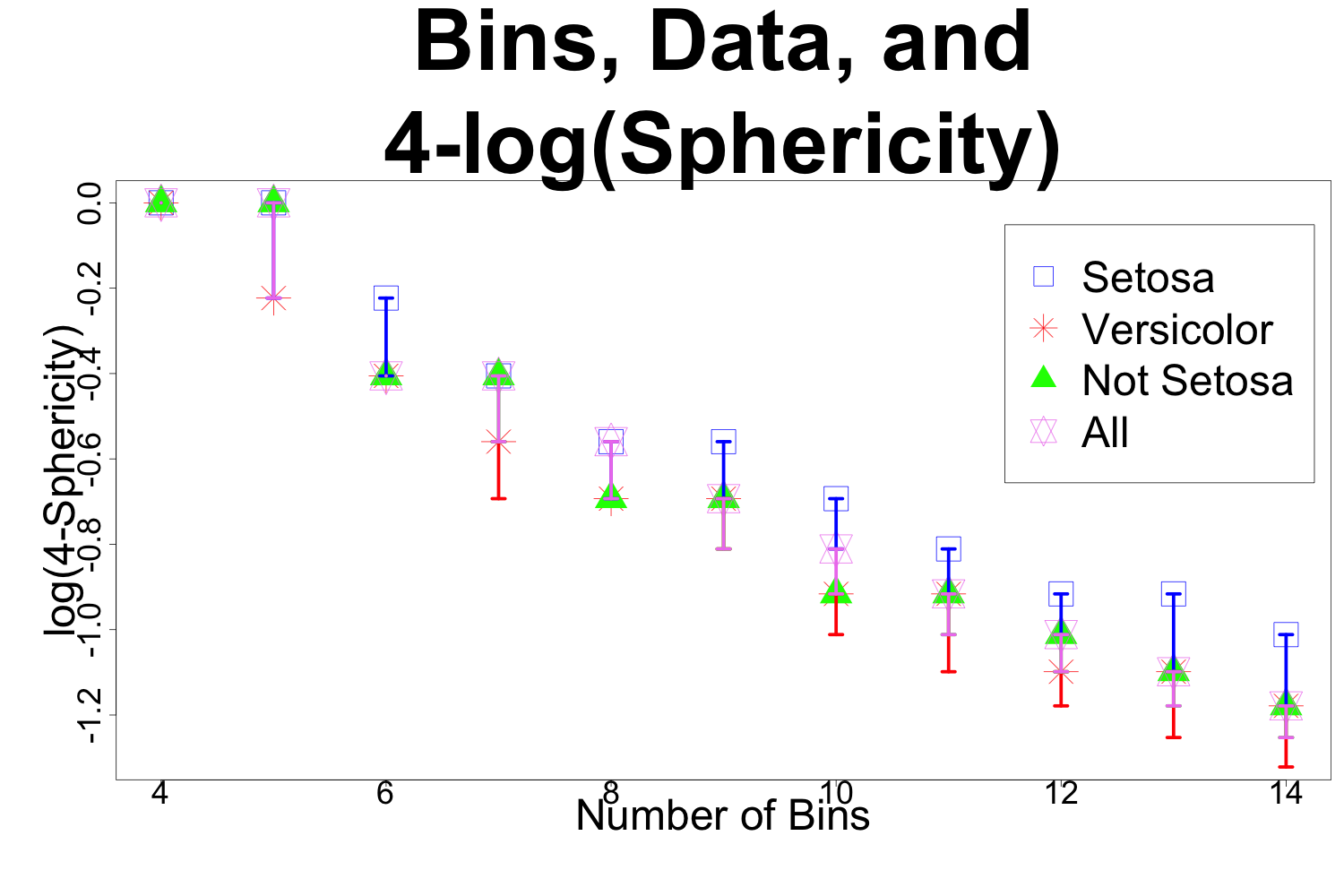}
    \caption{}
    \label{fig: iris_log_spher}
    \end{subfigure}
    \caption{\textbf{(a)} and \textbf{(b)} provide the bootstrapped estimates of the natural log of hyper-SP and sphericity, respectively.  The bootstrapped 95\% confidence intervals (CIs) are provided as bars surrounding the mean of each metric.}
    \label{fig: iris_results}
\end{figure*}

\begin{figure*}[h]
    \centering
    \begin{subfigure}[t]{.99\textwidth}
    \centering
    \includegraphics[width=0.90\linewidth]{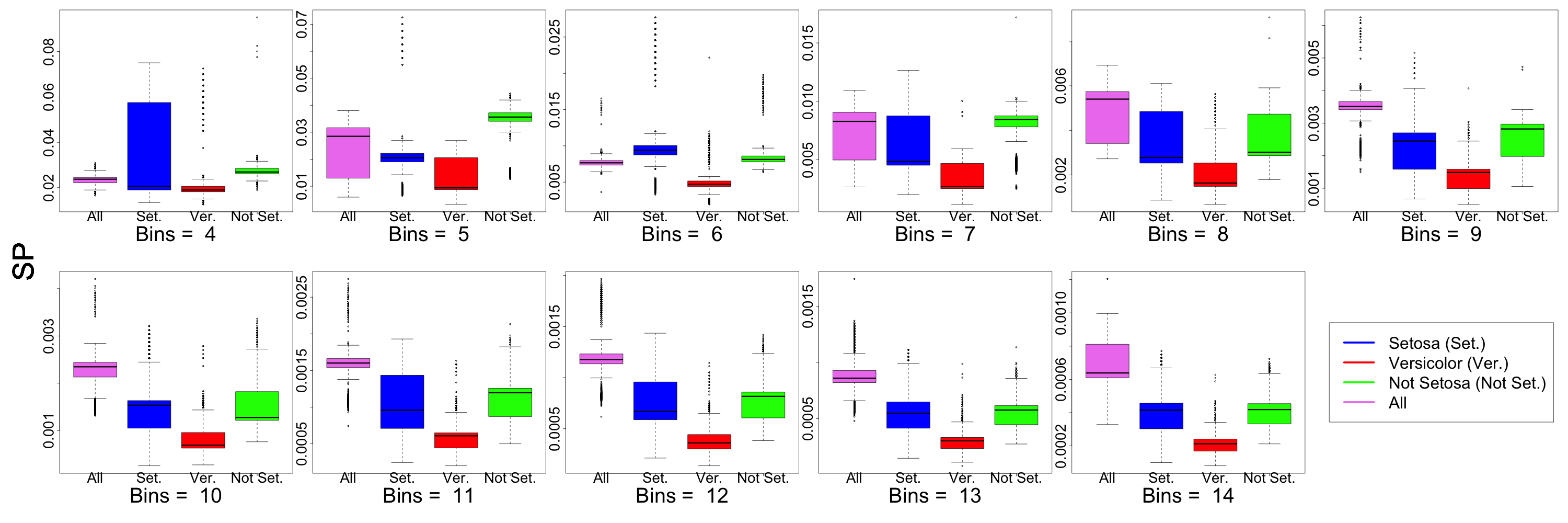}
    \caption{}
    \label{fig: iris_sp_box}
    \end{subfigure}\\
    \begin{subfigure}[t]{.99\textwidth}
    \centering
    \includegraphics[width=0.90\linewidth]{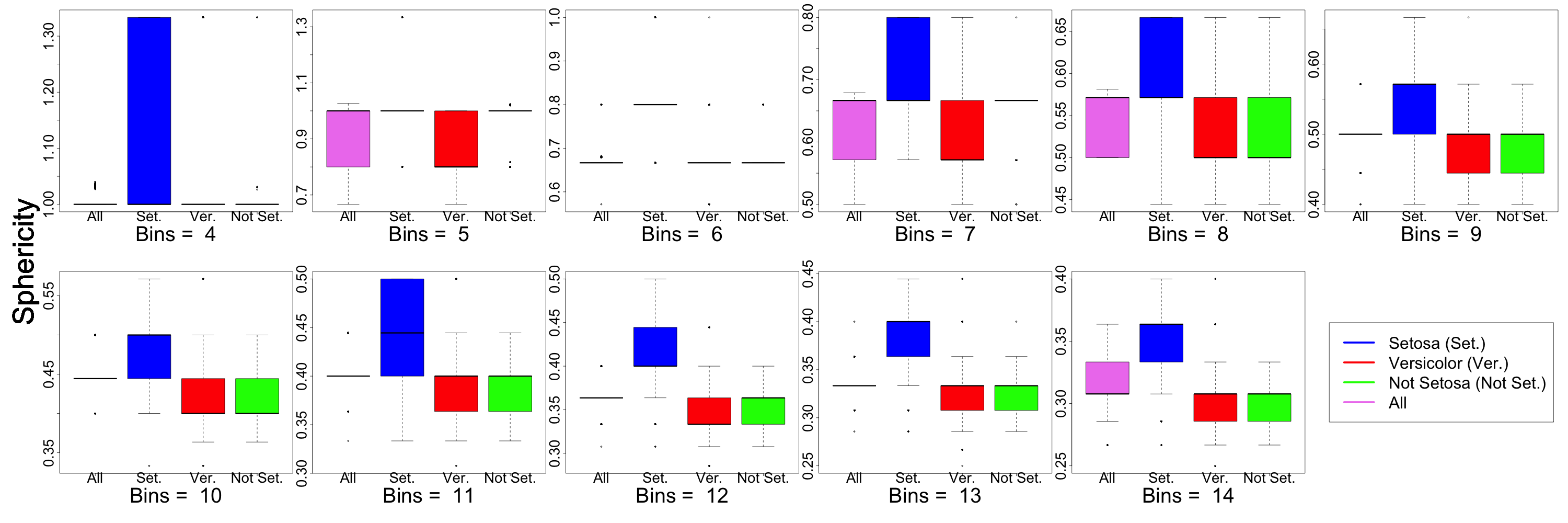}
    \caption{}
    \label{fig: iris_spher_box}
    \end{subfigure}
    \caption{Estimated values when using different bin sizes across different dimensions for \textbf{(a)} Hyper-Sphericity, \textbf{(b)} Hyper-SP, and the \textbf{(c)} Natural log of Hyper-SP.}
    \label{fig: iris_boxplots}
\end{figure*}

We wanted to observe if hyper-sphericity and hyper-SP would be able to distinguish between different subsets of the Iris data.  The four subsets were based upon the three Iris species: setosa, veriscolor, veriscolor and virginica (not setosa), and all three (all). We utilized the bootstrap to obtain 95\% confidence intervals (CIs) for both shape metrics for each of the four subsets.  After extracting the metrics in Python, we then analyzed and plotted our results in R (Figures \ref{fig: iris_results} and \ref{fig: iris_boxplots} and Table \ref{tab: iris_results}.  

\begin{figure*}[h]
    \centering
    \includegraphics[width=0.90\linewidth]{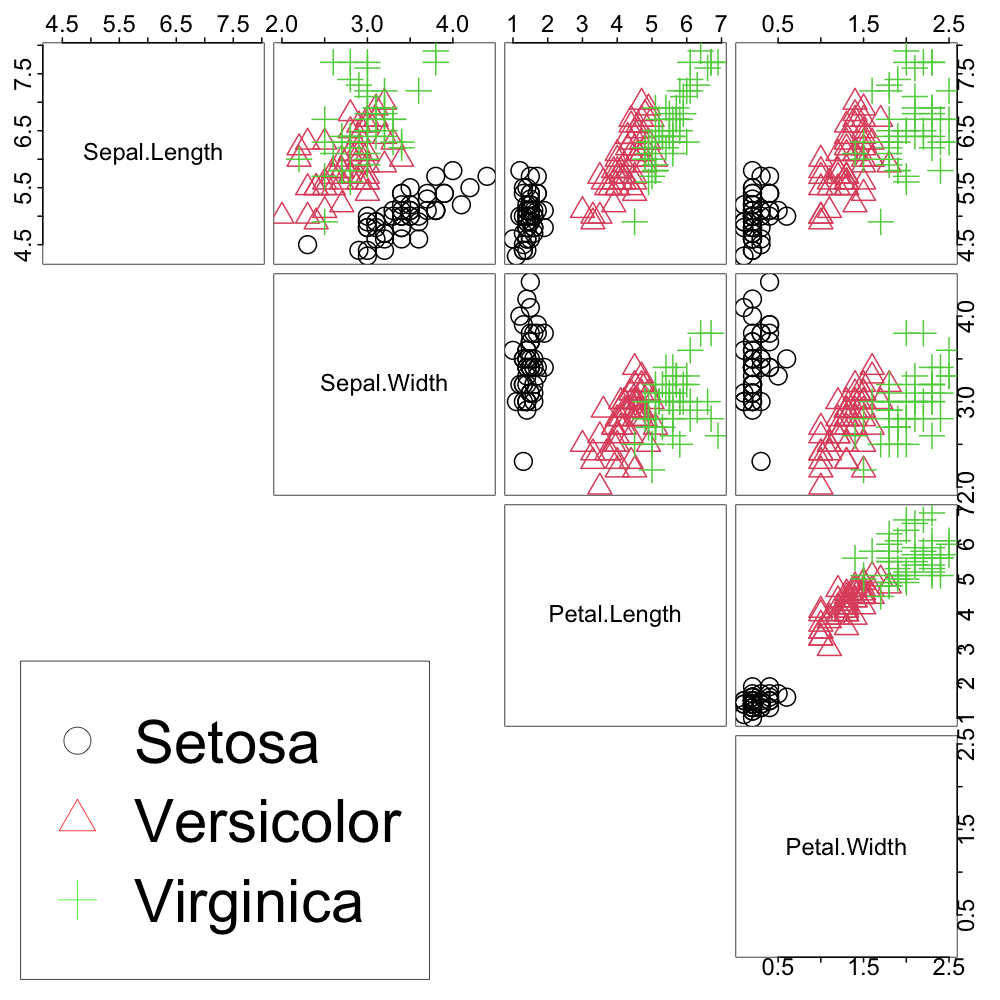}
    \caption{Scatterplot matrix of the Iris data. }
    \label{fig: iris_summary}
\end{figure*}

\begin{table*}[ht]
\centering
\caption{The bootstrapped mean and 95\% confidence intervals (CIs) for each of the shape metrics across the four different subsets and bin choices.}
\label{tab: iris_results}
\begin{tabular}{cc|ccc|ccc}
  \hline
 Class & Bins & SP 2.75\% & SP Median & SP 97.5\% & Spher 2.75\% & Spher Median & Spher 97.5\% \\
  \hline
  Setosa     & 4 & 0.017 & 0.021 & 0.065 & 1.000 & 1.000 & 1.000 \\
  Setosa     & 5& 0.008 & 0.021 & 0.025 & 0.800 & 1.000 & 1.000 \\
  Setosa     & 6& 0.005 & 0.010 & 0.025 & 0.667 & 0.800 & 0.800 \\
  Setosa     & 7& 0.004 & 0.005 & 0.010 & 0.667 & 0.667 & 0.667 \\
  Setosa     & 8 & 0.001 & 0.003 & 0.005 & 0.500 & 0.571 & 0.571 \\
  Setosa     & 9 & 0.001 & 0.003 & 0.003 & 0.485 & 0.571 & 0.571 \\
  Setosa     & 10 & 0.001 & 0.001 & 0.003 & 0.444 & 0.500 & 0.500 \\
  Setosa     & 11 & 0.001 & 0.001 & 0.002 & 0.390 & 0.444 & 0.444 \\
  Setosa     & 12 & 0.000 & 0.001 & 0.001 & 0.364 & 0.400 & 0.400 \\
  Setosa     & 13 & 0.000 & 0.001 & 0.001 & 0.333 & 0.400 & 0.400 \\
  Setosa     & 14 & 0.000 & 0.000 & 0.001 & 0.308 & 0.364 & 0.364 \\ \hline 
  Versicolor & 4 & 0.016 & 0.019 & 0.039 & 1.000 & 1.000 & 1.000 \\
  Versicolor & 5 & 0.007 & 0.009 & 0.024 & 0.800 & 0.800 & 0.800 \\
  Versicolor & 6 & 0.004 & 0.005 & 0.010 & 0.667 & 0.667 & 0.667 \\
  Versicolor & 7 & 0.002 & 0.003 & 0.005 & 0.500 & 0.571 & 0.571 \\
  Versicolor & 8 & 0.001 & 0.002 & 0.003 & 0.500 & 0.500 & 0.500 \\
  Versicolor & 9 & 0.001 & 0.001 & 0.002 & 0.444 & 0.500 & 0.500 \\
  Versicolor & 10 & 0.001 & 0.001 & 0.001 & 0.390 & 0.400 & 0.400 \\
  Versicolor & 11 & 0.000 & 0.001 & 0.001 & 0.333 & 0.400 & 0.400 \\
  Versicolor & 12 & 0.000 & 0.000 & 0.001 & 0.308 & 0.364 & 0.364 \\
  Versicolor & 13 & 0.000 & 0.000 & 0.000 & 0.286 & 0.333 & 0.333 \\
  Versicolor & 14 & 0.000 & 0.000 & 0.000 & 0.267 & 0.308 & 0.308 \\ \hline 
  Not Setosa & 4 & 0.024 & 0.028 & 0.030 & 1.000 & 1.000 & 1.000 \\
  Not Setosa & 5 & 0.031 & 0.036 & 0.040 & 1.000 & 1.000 & 1.000 \\
  Not Setosa & 6 & 0.007 & 0.008 & 0.013 & 0.667 & 0.667 & 0.667 \\
  Not Setosa & 7 & 0.004 & 0.008 & 0.009 & 0.571 & 0.667 & 0.667 \\
  Not Setosa & 8 & 0.003 & 0.003 & 0.005 & 0.500 & 0.500 & 0.500 \\
  Not Setosa & 9 & 0.002 & 0.003 & 0.003 & 0.444 & 0.500 & 0.500 \\
  Not Setosa & 10 & 0.001 & 0.001 & 0.002 & 0.400 & 0.400 & 0.400 \\
  Not Setosa & 11 & 0.001 & 0.001 & 0.001 & 0.364 & 0.400 & 0.400 \\
  Not Setosa & 12 & 0.001 & 0.001 & 0.001 & 0.333 & 0.364 & 0.364 \\
  Not Setosa & 13 & 0.000 & 0.001 & 0.001 & 0.308 & 0.333 & 0.333 \\
  Not Setosa & 14 & 0.000 & 0.000 & 0.001 & 0.267 & 0.308 & 0.308 \\ \hline 
  All        & 4 & 0.020 & 0.023 & 0.027 & 1.000 & 1.000 & 1.000 \\
  All        & 5 & 0.011 & 0.028 & 0.034 & 0.800 & 1.000 & 1.000 \\
  All        & 6 & 0.007 & 0.008 & 0.009 & 0.667 & 0.667 & 0.667 \\
  All        & 7 & 0.005 & 0.008 & 0.010 & 0.571 & 0.667 & 0.667 \\
  All        & 8 & 0.003 & 0.005 & 0.006 & 0.500 & 0.571 & 0.571 \\
  All        & 9 & 0.002 & 0.004 & 0.004 & 0.444 & 0.500 & 0.500 \\
  All        & 10 & 0.001 & 0.002 & 0.003 & 0.400 & 0.444 & 0.444 \\
  All        & 11 & 0.001 & 0.002 & 0.002 & 0.364 & 0.400 & 0.400 \\
  All        & 12 & 0.001 & 0.001 & 0.002 & 0.333 & 0.364 & 0.364 \\
  All        & 13 & 0.001 & 0.001 & 0.001 & 0.308 & 0.333 & 0.333 \\
  All        & 14 & 0.000 & 0.001 & 0.001 & 0.286 & 0.308 & 0.308 \\
   \hline
\end{tabular}
\end{table*}

\section{Discussion}

We have shown that many common objects have unique values for hyper-SP and hyper-sphericity.  This provides evidence that these shape metrics have unique properties for multidimensional objects.  Thus, obscure objects in $n$D space can be compared and contrasted against their known counterparts. 

Our algorithm for computing hyper-SP and hyper-sphericity for continuous data or objects depends upon multidimensional histograms.  These multidimensional histograms convert continuous space into discrete space.  However, the choice of the number of bins changes the final estimated value of our metrics (Fig. \ref{fig: uniform}).  Further, the number of dimensions also has an effect on our algorithms estimated value.  Thus, further refinements can be made to the computation of hyper-sphericity and hyper-SP.  Utilizing specialized histogram estimation techniques may offer improved results.  

We showed that our metric can also be applied to observational data by performing experiments using the Iris dataset.  We again observed that the number of bins affected the estimated value of hyper-SP and hyper-sphericity (Table \ref{tab: iris_results}, Fig. \ref{fig: iris_log_sp} - \ref{fig: iris_log_spher}, and Fig. \ref{fig: iris_boxplots}).  Our experiments also confirmed that for 4 dimensions, utilizing 6 bins was the optimal choice for hyper-SP (Figs. \ref{fig: hyper_sp}, \ref{fig: log_hyper_sp}, \ref{fig: iris_log_sp}, \ref{fig: iris_sp_box}).  While our theoretical results suggested that 4 bins was optimal when $n=4$ (Fig. \ref{fig: hyper_spher}), our experiment also suggested that 6 bins was a consistent choice as well (Fig. \ref{fig: iris_spher_box}.  However, both metrics had overlapping bootstrapped 95\% CIs across the different subsets and dimensions (Tab. \ref{tab: iris_results}, Fig. \ref{fig: iris_log_sp}, and Fig. \ref{fig: iris_log_spher}).  When the bins was 6, the Setosa subset's CI for  hyper-sphericity (0.667, 0.800) did just overlap with the other subsets (0.667, 0.667).  This does provide evidence that the Setosa class is marginally significant.  Conversely, hyper-SP had overlapping CI's across the subsets.  This suggests that the shape of the data is not changing substantially.  

The bootstrapped distributions of our metrics do not remain consistent for all of the subsets for any given bin size (Fig. \ref{fig: iris_boxplots}).  If the data was homogeneous, we would expect each subset to provide similarly shaped distributions.  For example, if the median would be in the same general location of the boxplot but the entire boxplot was shifted and still overlapping, this would suggest that the subsets are similar (Fig. \ref{fig: iris_spher_box}'s Bins=10 Ver. and Not Set.).  However, in nearly all of the panels, there is at least one boxplot that differs from the others (Fig. \ref{fig: iris_boxplots}).  This suggests that the data is less homogeneous than the numerical values suggest by themselves.  Thus, there is some evidence that the data's shape does change when using different subsets of the data.  This suggests that future experiments should capture these different subsets to ensure that future experiments capture all of the diversity in the population of interest.  

This foundational work provides a framework for analyzing data as images.  This helps to connect the fields of Geometry with Image Analysis (or Computer Vision) and Data Science by describing the shape of multidimensional data.  Future applications of using shape metrics to describe data include outlier detection and data homogeneity.  

\section{Conclusions}   

Both hyper-SP and hyper-Sphericity provide ways of summarizing $n$D shapes and have theoretical values for some known objects.  We have also shown that these shape metrics can be extended to contexts beyond Geometry and the to fields of Image Analysis and Data Science by converting data into multidimensional images.  This foundational work provides evidence that future shape metrics can be applied for the analysis and characterization of multidimensional data.  

\section{Acknowledgements}

UVA Engineering Graduate Writing Lab Peer Review Group provided valuable feedback during initial drafts of this manuscript.

We would also like to thank the Zang Lab for Computational Biology at the University of Virginia for their support. 

\clearpage

\bibliographystyle{IEEEtran}
\bibliography{Zotero.bib}

\end{document}